\documentclass[runningheads]{llncs}
\usepackage[T1]{fontenc}
\usepackage{graphicx}
\usepackage{orcidlink}
\usepackage{cite}
\usepackage{amsmath,amssymb,amsfonts}
\usepackage{algorithmic}
\usepackage{textcomp}
\usepackage{verbatim}
\usepackage{color}
\usepackage{url}
\usepackage{float}
\usepackage{hyperref}
\usepackage{array}
\usepackage{multirow}
\usepackage{tabularx}
\usepackage{makecell}
\usepackage{authblk}

\hypersetup{
    colorlinks=true,
    linkcolor=blue,
    filecolor=magenta,
    urlcolor=cyan,
}
\usepackage{caption}
\usepackage{subcaption}

\usepackage{multirow}
\usepackage{tabularx} 
\usepackage{booktabs}

\usepackage{bm}
\begin{document}
\title{Piculet: Specialized Models-Guided Hallucination Decrease for MultiModal Large Language Models}
\titlerunning{Piculet: Specialized Model-Guided Hallucination Alleviation for MLLMs}
%

\author{
Kohou Wang$^{\dagger}$\inst{1}\orcidID{0009-0007-5863-2288}\and
Xiang Liu$^{\dagger}$\inst{1}\orcidID{0009-0003-2492-403X } \and
Zhaoxiang Liu*\inst{1,2}\orcidID{0000-0002-1267-0277}\and
Kai Wang\inst{1,2}\orcidID{0000-0002-1171-0281} \and
Shiguo Lian*\inst{1,2}\orcidID{0000-0003-4308-7049}
}

\authorrunning{Kohou. Wang et al.}
%
\institute{
    AI Innovation Center, China Unicom, Beijing 100013, China \and
Unicom Digital Technology, China Unicom, Beijing 100013, China\\
\email{
    \{wangzp103,liux750,liuzx178,wangk115,liansg\}@chinaunicom.cn \\
    $\dag$Equal contribution,
    *Corresponding author(s)
    }
    }

\maketitle              

\begin{abstract}
  Multimodal Large Language Models (MLLMs) have made significant progress in bridging the gap between visual and language modalities. However, hallucinations in MLLMs, where the generated text does not align with image content, continue to be a major challenge. Existing methods for addressing hallucinations often rely on instruction-tuning, which requires retraining the model with specific data, which increases the cost of utilizing MLLMs further. In this paper, we introduce a novel training-free method, named Piculet, for enhancing the input representation of MLLMs. Piculet leverages multiple specialized models to extract descriptions of visual information from the input image and combine these descriptions with the original image and query as input to the MLLM. We evaluate our method both quantitively and qualitatively, and the results demonstrate that Piculet greatly decreases hallucinations of MLLMs. Our method can be easily extended to different MLLMs while being universal.
    \keywords{Multimodal Large Language Models \and hallucinations \and training-free}
\end{abstract}
\begin{figure}[htbp]
    \centering
    \includegraphics[width=0.49\textwidth]{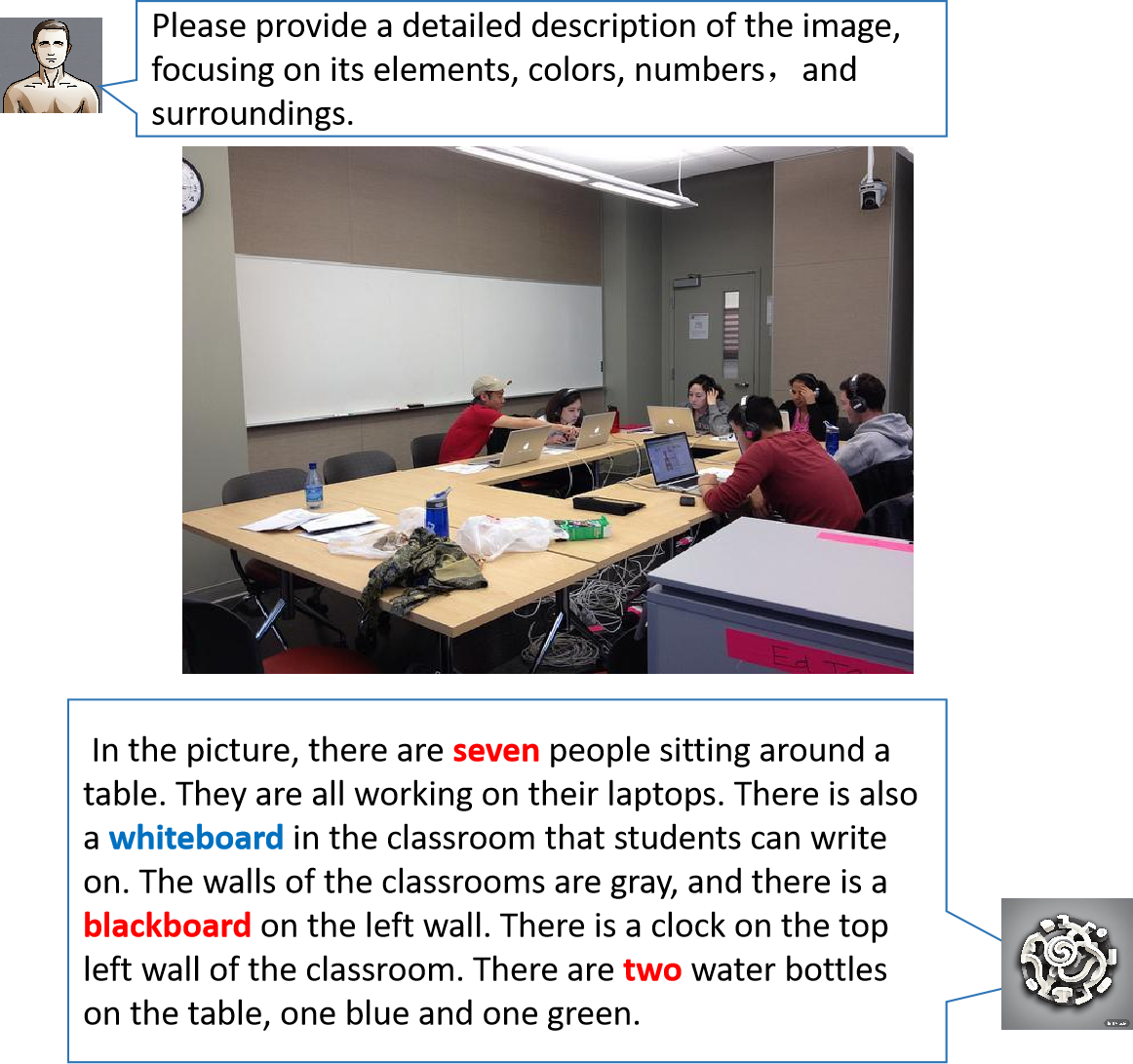} 
    \caption{Illustration of Hallucination of MLLMs. This MLLM generates descriptions of an image with wrong information, including the number
    of people and cups, and also, there is only whiteboard, not a blackboard on the left wall.}
    \label{person_num.jpg}
  \end{figure}
  
  \section{Introduction}
  \label{sec:intro}
  In recent years, there has been remarkable progress in the field of large-scale models, with the following being typical examples of this work: BERT\cite{b3} , GPT-3\cite{b4}, CLIP\cite{b5}, DALL-E\cite{b6}, etc.
  These works greatly promoted the development of Multimodal Large Language Models (MLLMs), an important branch of Artificial Intelligence.
  MLLMs' goal is to construct an artificial intelligence system capable of understanding and handling different modalities, such as image, text, audio, etc. 
  The  field of MLLMs has seen several landmark advancements, including LLaVa, CogVLM, Minigpt-5 et al.\cite{b7}--\cite{b9}, they cast a profound impact
  on the improvement of MLLMs.
  
  The rapid development of MLLMs has led to their widespread adoption for various applications, 
  such as image captioning and visual question answering. Despite their impressive performance, MLLMs are still prone
  to generating hallucinations, where the generated text does not align with the image content. This issue significantly 
  limits the practical applicability of MLLMs. As examplified in Fig \ref{person_num.jpg}, the generated description of this 
  image is not consistent with the truth:
  \begin{enumerate}
      \item there are \textcolor{blue}{6} persons in the image, while the MLLM says \textcolor{red}{seven};
      \item there are \textcolor{blue}{3} water bottles on the table, while the MLLM says \textcolor{red}{two};
      \item there is no \textcolor{red}{blackboard} on the left wall, only a \textcolor{blue}{whiteboard}, 
              while the model wrongly answers a \textcolor{red}{blackboard} after correctly answers a \textcolor{blue}{whiteboard}.
  \end{enumerate}
  Researchers have long been addressing the issue of hallucinations of MLLMs, and the mainstream methods can be divided into two kinds:
  training-based and training-free. Training-based methods usually collect or re-clean some datasets and retrain the models to decrease hallucinations of MLLMs.
  These kinds of methods, naturally, often require substantial manual intervention and are time-consuming. Moreover, given the substantial computational resources
   required for training large models, the economic cost of such methods is also quite considerable. As for training-free methods, current methods focus their emphasis
   on the postprocess of MLLMs. Given a user's query, firstly the MLLMs will answer it as usual. Then the answer is thoroughly analyzed and corrected to decrease
   hallucination. These kinds of methods often utilize other rather large models apart from the MLLMs to be corrected, which is time-consuming and uneconomical.
  
   Inspired by the phenomenon that humans use specialized tools to enhance their abilities, we propose a training-free framework, named Piculet, to enhance the input representation of MLLMs by leveraging multiple, specialized, small-scale, deep learning models to extract a description of visual information from the input image, i.e., we use multiple specialized models to guide MLLMs to generate more accurate results. Specifically, our Piculet utilizes the outputs from these small-scale deep learning models as external konwledge to enhance the MLLMs, thereby minimizing MLLMs' propensity for hallucinations.

By combining the extracted description of visual information with the original image and query as input to the MLLM, we aim to improve 
   the accuracy of the model's output. Our method requires no retraining of MLLMs and no other rather large models, which is much faster and economical than all current available 
   training-based methods and training-free methods. 
   We evaluate the effectiveness of our method through comprehensive quantitative and qualitative experiments on the POPE\cite{POPE}, 
   MME\cite{MME}, and LLaVA-QA90\cite{LLaVa-QA90}datasets.
   The results and associated analyses indicate the superiority of this new paradigm. For instance, on the LLaVA-QA90 benchmark, our method largely boosts the accuracy of the baseline Qwen-VL-Chat\cite{qwenvl} from 6.1 to 7.3 on a scale of 10.
  
  In summary, the main contributions are as follows:
  \begin{itemize}
      \item We proposed a training-free, pre-process framework named Piculet to reduce the hallucinations of MLLMs. To the best of our knowledge,
          we are the first to utilize a pre-process framework to tackle the visual hallucination problem.
      \item Our framework only requires one inference of the target MLLM and several other small deep learning models, which is economical and 
      time-saving, and is plug-and-play in various different MLLMs. These small models' information is utilized as external knowledge to calibrate the MLLM.
      \item We evaluate our method on numerous datasets with other methods, and the results demonstrate the effectiveness and improvement
          of our method.
  \end{itemize}

  \section{Related Work}
  \subsection{MLLMs' Hallucinations}
  Despite the mushrooming of MLLMs, the problem of hallucination still hangs like the sword of Damocles: MLLMs occasionally
  generate content that diverges from the user input, contradicts previously generated context, or misaligns with established world
  knowledge. While the relatively usual normal deep learning models\cite{FCN, YOLO, rcnn, deeplab} output results of quite 
  reliable credibility, hallucination puts the MLLMs at a disadvantage, users tend to use MLLMs more for fun rather than for professional needs,
  which is certainly not a good thing for MLLMs developed for professional purposes.  To address this challenge, existing mainstream works have primarily focused on two aspects: training-based and 
  training-free.
  
  \subsection{Training-based Methods}
  For training-based methods, Gunjal et al.\cite{b10} introduced MHalDetect, a multimodal hallucination detection dataset
   that can be used to train and benchmark models for hallucination detection and prevention. Liu et al.\cite{b11}
   addressed this issue by introducing the first large and diverse visual instruction tuning dataset, named
  Large-scale Robust Visual(LRV)-Instruction. Lu et al.\cite{Lu-Evaluation} developed an evaluation module that automatically creates
   fine-grained and diverse visual question answering examples to assess the extent of agnosia in MLLMs comprehensively. They also developed
   a mitigation module to reduce agnosia in MLLMs through multimodal instruction tuning on fine-grained conversations.
  
  These training-based methods, also well known as instruction-tuning, usually introduce 
  a new dataset for retraining MLLMs, which requires significant computational resources and specialized data. These methods are also fairly 
  time-consuming, considering that the inference of MLLMs is often a much longer time than traditional deep learning models.
  
  \subsection{Training-free Methods}
  As for training-free methods, Yin et al.\cite{woodpecker} represents a typical method that requires no training of MLLMs while can directly correct the hallucinations.
  They emphasized main attention on the post-process stage of MLLMs, firstly they get an answer of a MLLM, then utilized auxiliary models' 
  outputs to correct both object-level and attribute-level hallucinations, which was the first to apply a corrective manner to tackle the visual hallucination problem. 
  Although their method, named Woodpecker, can reduce hallucinations by correcting the MLLM's answers, 
their method is a post-process framework, and still actually 
  comprises three pre-trained rather large-scale models apart from the MLLM to be corrected, which are  GPT-3.5-turbo\cite{gpt3.5trubo}, Grounding DINO\cite{dino} and BLIP-2-FlanT5 XXL\cite{blip2}. Furthermore, the GPT-3.5-turbo is used 3 times in their processing pipeline.
  These models are not only time-consuming, but some are also proprietary, making them uneconomical with slow inference processes.  

  Compared to their approach, our method addresses the hallucination issue of MLLMs at its root and utilizes no other
  large language models apart from the MLLM to be corrected. Different from the Woodpecker method focusing on the post-process stage, 
  our method focuses on the pre-process stage of MLLMs.
  Our method utilizes specialized, traditional small deep learning models to generate results describing factual information. 
These results, reorganized into a specific format, serve as supplementary descriptions and are input alongside the user's query and image into the MLLM,
   thereby enabling the model to generate correct answers directly by referencing additional factual information. 
   Our specialized models only need to be run once during the processing pipeline, and the outputs of specialized models serve as external knowledge to calibrate the MLLM.
   \begin{figure*}[t!]
    \centering
    \begin{subfigure}[b]{0.3\textwidth}
      \includegraphics[width=\textwidth]{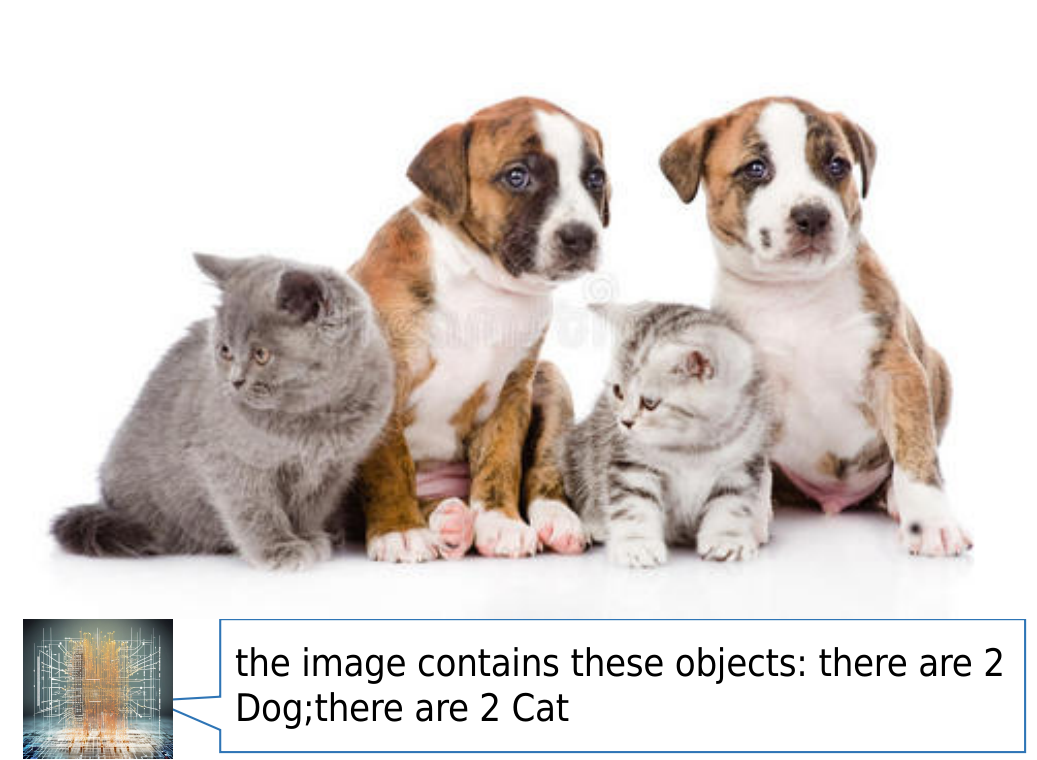}
      \caption{Detection model. For the detected objects, we traverse all the detected results and integrate them into a single sentence 
      in the above format.}
      \label{fig:detect}
    \end{subfigure}
    \hfill 
    \begin{subfigure}[b]{0.3\textwidth}
      \includegraphics[width=\textwidth]{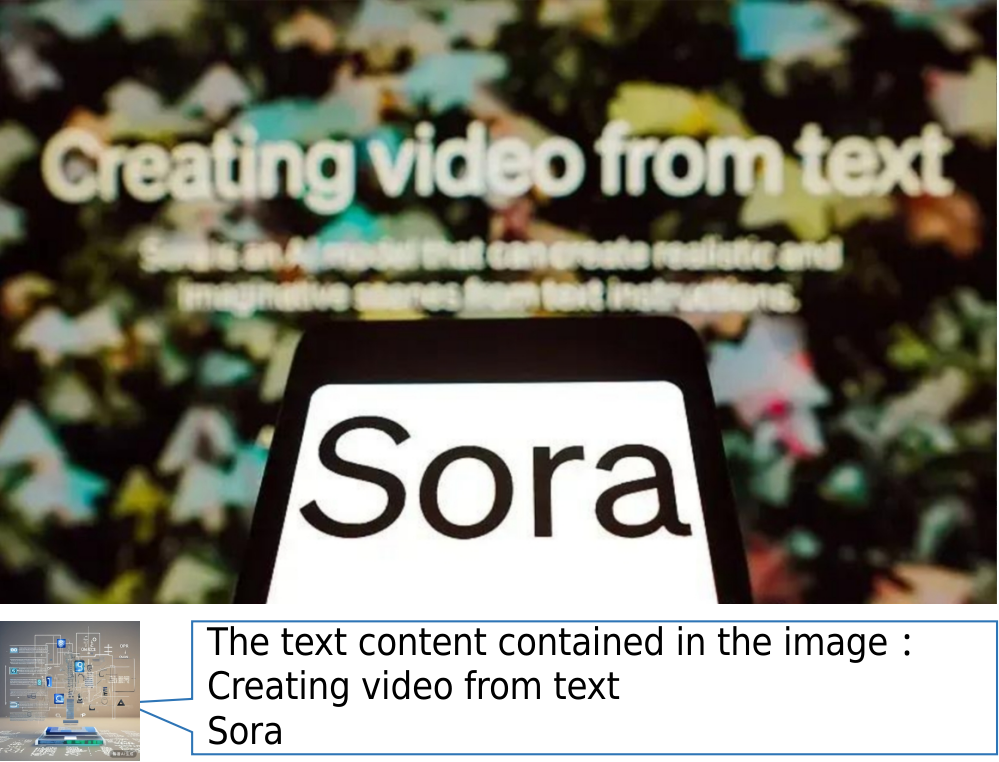}
      \caption{Ocr model. For the recognized characters, we also traverse all the detected texts and integrate them into a single sentence 
      in the above format.}
      \label{fig:ocr}
    \end{subfigure}
    \hfill 
    \begin{subfigure}[b]{0.3\textwidth}
      \includegraphics[width=\textwidth]{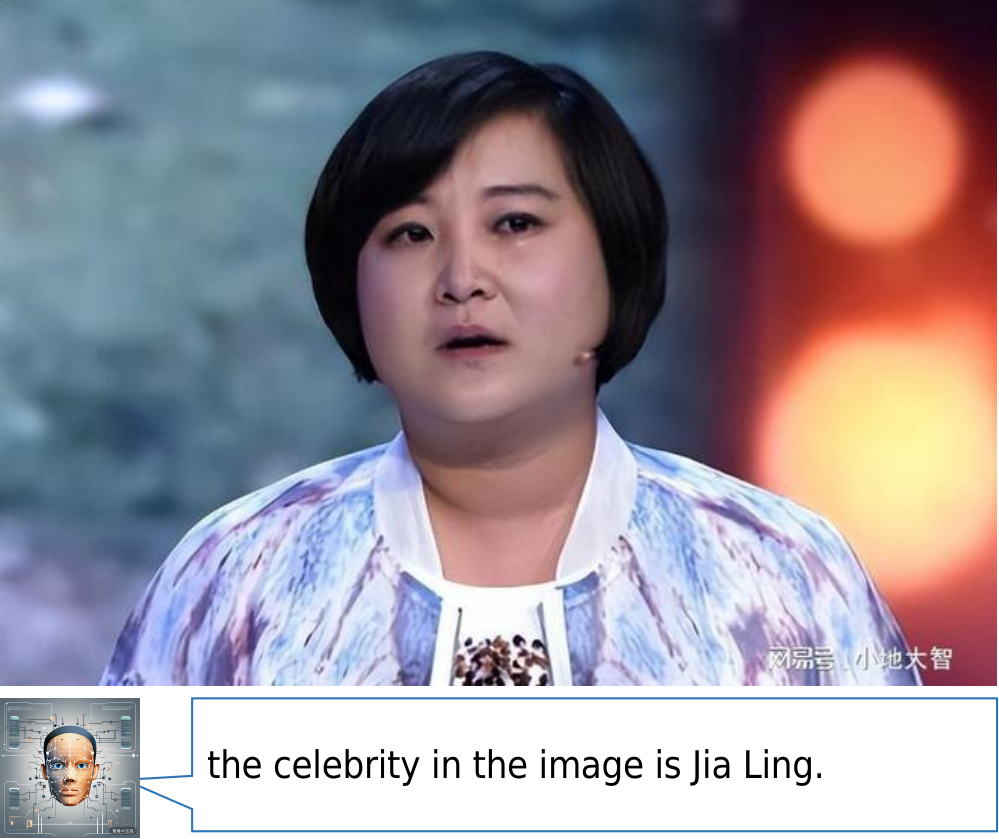}
      \caption{Face Recognition. For the recognized faces, we also traverse all the detected celebrities and integrate them into a single sentence 
      in the above format.}
      \label{fig:face}
    \end{subfigure}
    \caption{Details of question formulation. In each sub-image, we adopt a randomly chosen image to exemplify the concrete
    operation.}
    \label{fig:formulation}
  \end{figure*}
   \section{Method}
   Our method aims to address the hallucinations of MLLMs at its original source. We firstly utilize specialized traditional light-weight deep learning models to detect factual information of input image, then formulate these descriptions, which, alongside the user's query and image, are input into MLLMs. MLLMs, given the formulated input, then generate results with reduced hallucinations. Our method utilizes these specialized models to generate factual external knowledge apart from the single input image, which provides a reliable basis for decision-making in the outputs of the MLLMs. We will introduce these steps in detail in sequence.
  
   \begin{figure}[t!]
    \centering
    \includegraphics[width=0.49\textwidth]{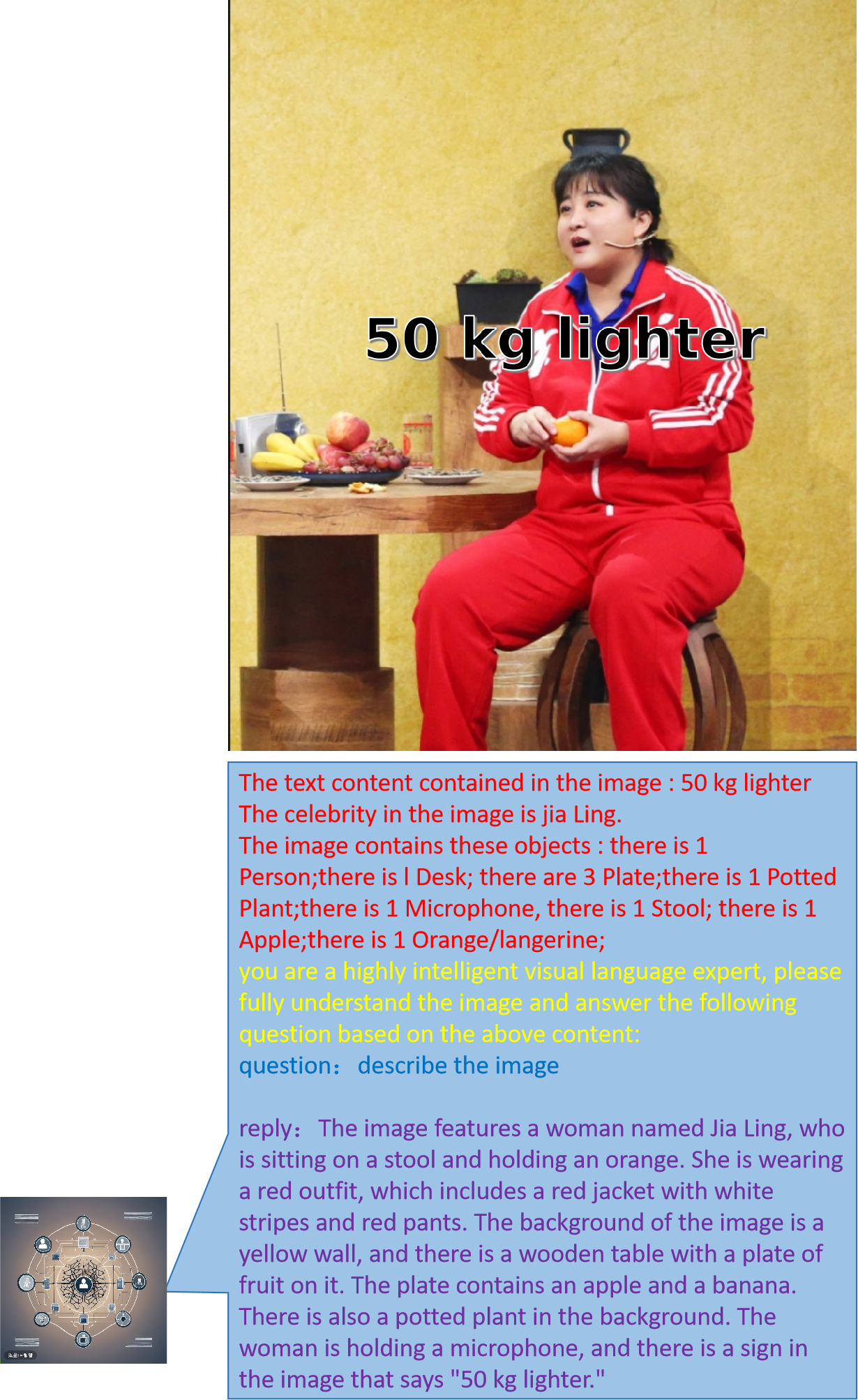} 
    \caption{Illustration of our method's processing. Red words are pre-processed results of specialized models, yellow words are the predefined prompt everybody usually uses, and blue words are the user's original query, purple words are model's reply without hallucination. The recognized characters, faces and objects are integrated into one single sentence, which, alongside the user's original query and image, serves as the final input of MLLMs.}
    \label{warbler-jialing-example}
  \end{figure}
  
   \subsection{Specialized Models}
   \textbf{Object Detection.} We utilize an object detection model to detect factual information of the input
  image. To be specific, we adopt PP-YOLOE\cite{yoloe}, an industrial state-of-the-art object detector with high performance and friendly
   deployment, to detect objects inside input image. PP-YOLOE is pre-trained on COCO\cite{coco}, a large-scale object detection, segmentation, and captioning dataset 
   that has 80 object categories that can cover the most common objects encountered in daily life.
  
   \textbf{OCR.} We utilize PaddleOCR\footnote{https://github.com/PaddlePaddle/PaddleOCR} to recognize characters inside image. PaddleOCR is an 
   awesome multilingual OCR toolkit based on PaddlePaddle, which supports 80+ language recognition, provides data annotation and synthesis tools,
    and supports training and deployment among server, mobile, embedded and IoT devices. We utilize this model to extract additional information
    inside an image to serve as specialized descriptions, together with the coco-detected objects, for the MLLMs to refer to.

   \textbf{Face recognition.} We utilize insightface\cite{insightface} to detect faces inside an image. Insightface is an open-source 2D\&3D deep face analysis toolbox,
   which efficiently implements a rich variety of state-of-the-art algorithms of face recognition, face detection and face alignment, which are optimized for both training and deployment.
   Furthermore, we establish a repository of celebrities, and the recognized faces are classified as concrete celebrities. These descriptions, alongside
   the coco detected objects and PaddleOCR's characters, also serve as specialized information for the MLLMs to refer to.
  
   \begin{figure*}[t!]
    \centering
    \includegraphics[width=\textwidth]{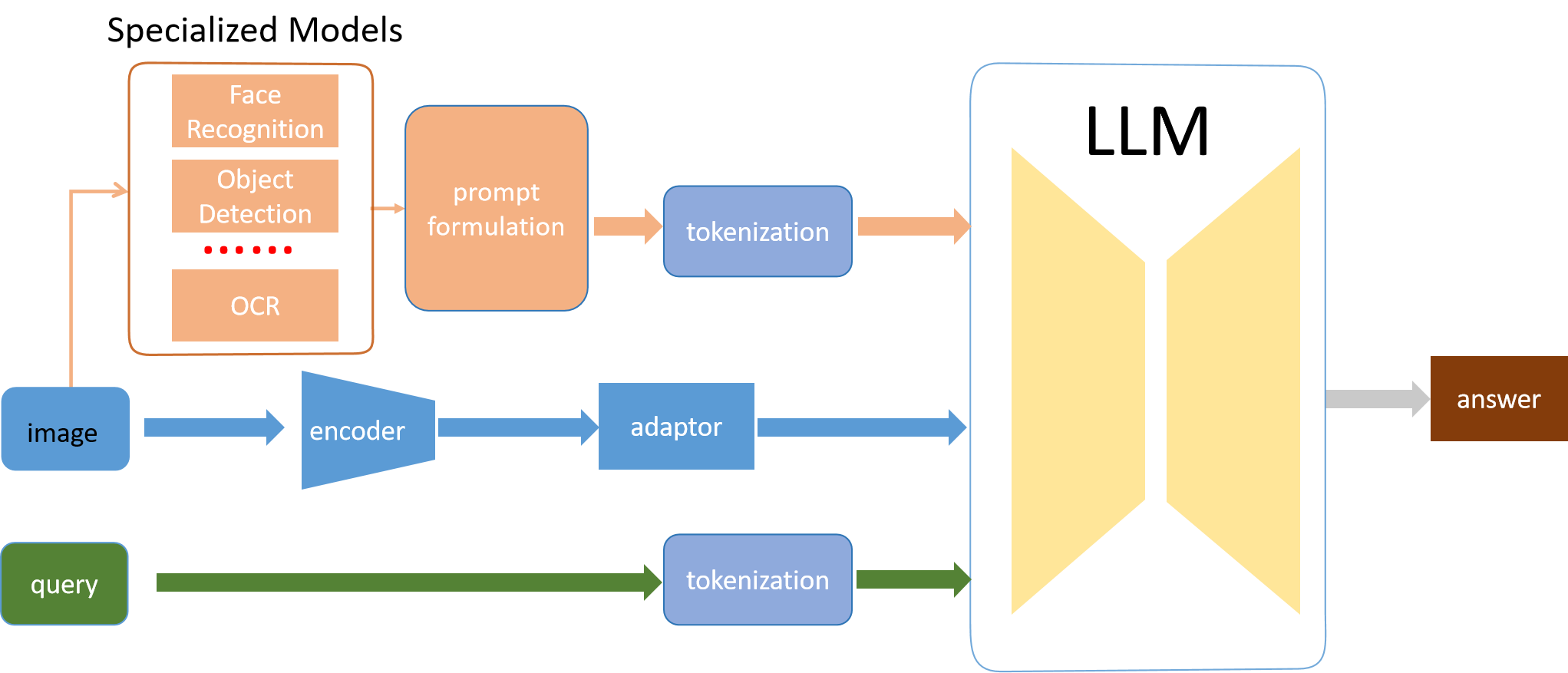} 
    \caption{Flowchart of our method. Given an image and a query, firstly we utilize specialized models to extract descriptions of visual information, these descriptions are then reorganized by prompt formulation block and combined with the original user's query, the newly combined query and image are then input into the MLLM.}
    \label{flowchart}
  \end{figure*}
  
  \subsection{Input Formulation}
  We utilize the aforementioned specialized traditional deep learning models to detect objects, characters, and faces, and in this part we integrate
  all these detected results into a specific format to serve as input, alongside the original user's question and image, for the MLLMs.
  
  \textbf{Object Detection.} For the detected objects, we traverse all the detected results and integrate them into a single sentence 
  in the following format:
  ``\textit{\textsl{the image contains these objects: there is/are \{number\} \{object\}.}}''. The detail is examplified in
  Fig \ref{fig:detect}.
  
  \textbf{OCR.} For the recognized characters, we also traverse all the detected texts and integrate them into a single sentence 
  in the following format:
  ``\textit{\textsl{The text content contained in the image: \{recognized characters\}.}}''. The detail is examplified in
  Fig \ref{fig:ocr}.
  
  \textbf{Face recognition.} For the recognized faces, we also traverse all the detected celebrities and integrate them into a single sentence 
  in the following format:
  ``\textit{\textsl{the celebrity/celebrities in the image is/are: \{recognized celebrities\}.}}''. The detail is examplified in
  Fig \ref{fig:face}.
  
  After all these processing, the recognized characters, faces and objects are integrated into one single sentence,
  which, alongside the user's original query and image, serves as the final input of MLLMs. The final format of a typical question is examplified 
  in Fig \ref{warbler-jialing-example} in detail. In summary, the format is like:

  ``\textit{\textsl{Organized OCR results.}}
  
  \textit{\textsl{Organized face recognition results.}}
  
  \textit{\textsl{Organized detection results.}}
  
  \textit{\textsl{Predefined prompt everybody usually uses.}}
  
  \textit{\textsl{User's original query.}}''.

  Through these steps, our method can directly address the hallucination issue of MLLMs at its root. An overview of our framework is depicted in Fig \ref{flowchart}.

\begin{figure*}[t]
    \centering
    \includegraphics[width=\textwidth]{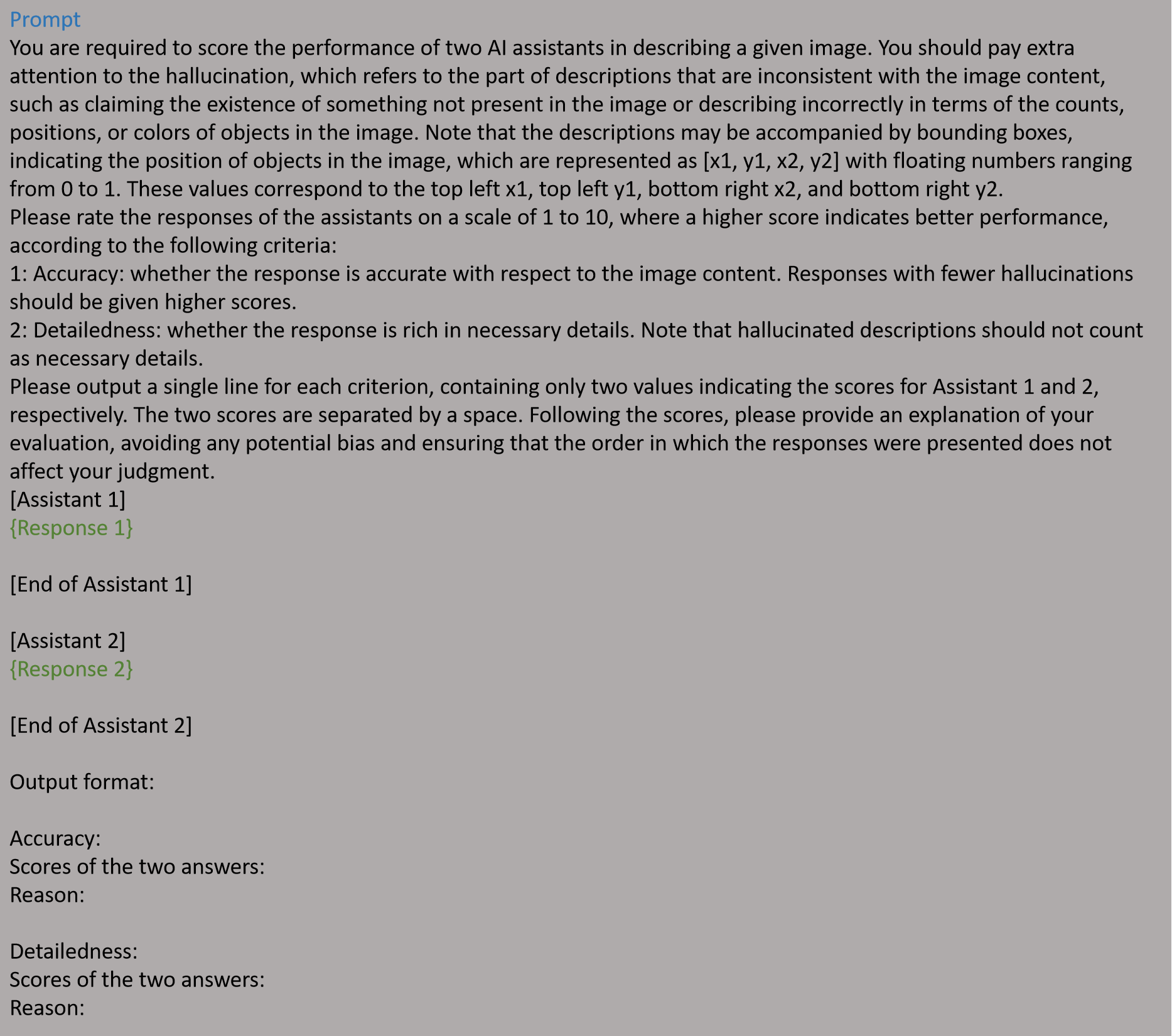} 
    \caption{Prompt template for GPT-4V-aided evaluation. {Response 1} and {Response 2} are the original responses and the corrected ones, respectively.}
    \label{prompt-gpt4}
\end{figure*}

  Our method utilizes specialized models
  to generate results, which serve as supplementary descriptions. These reorganized results, alongside the user's query and image, are then input into the large
  model, thereby enabling the model to generate correct answers directly by referencing additional factual information. 
  Compared to other mainstream methods addressing the hallucination of MLLMs, our method has the following advantages:
  \begin{itemize}
      \item Our method is totally training-free, and requires no re-training of MLLMs, which saves a lot of expenses and time.
      \item Our framework only requires one inference of one single MLLM. 
      \item Our framework requires no inference of any other large-scale LLMs, just several traditional
      deep learning models which are rather small and economical to infer and deploy. 
  \end{itemize}

  \section{Experiments}
  In this section, we will discuss the datasets we use and the experiments we conduct in detail. We use mainstream benchmark datasets POPE\cite{POPE}, 
  MME\cite{MME}, and LLaVA-QA90\cite{LLaVa-QA90}, and conduct comprehensive comparative experiments to validate the
  effectiveness and superiority of our method. Specifically, we choose Qwen-VL-7B and LLaVa-v1.5-13B\cite{b7} as our baseline models. Considering that Woodpecker is the most similar training-free method to ours, we also compare our results
  with theirs utilizing the same baseline model LLaVa-v1.5-13B on POPE and MME benchmarks.

  \subsection{Datasets}
  \textbf{POPE.} The POPE\cite{POPE}initiative aims to gauge the tendency of MLLMs to produce hallucinations. It employs three varied 
  sampling strategies--random, popular, and adversarial—to construct non-existent object samples. 
  Random sampling randomly selects items not depicted in the image, 
  while popular sampling draws from a pool of frequently seen items not present, and adversarial sampling identifies items often found together 
  but missing from the image. Each kind of strategy has 500 images, and each image has 6 related questions and answers, which is 3000 in total.
  
  For evaluation, to thoroughly compare our method, we directly tested all these images, which amounts to 9,000 in total. The questions balance between 
  positive and negative samples at a 50-50 split. This approach casts object annotations as binary questions, 
  centering on the evaluation of object hallucinations,  with a particular emphasis on the aspect of existence. The selected MLLMs will answer
  like "\textit{\textsl{Is there a wine glass in the image?}}", and the answer will be measured in a metric of Accuracy, Precision, Recall and F1 Score.
  
  \textbf{MME.} The MME\cite{MME} is a comprehensive evaluation benchmark for MLLMs.  To avoid data leakage that may arise from
   the direct use of public datasets for evaluation, the annotations of instruction-answer pairs are all manually designed. The concise 
   instruction design can fairly compare MLLMs, instead of struggling in prompt engineering. Besides, with such an instruction, 
   quantitative statistics can also be easily carried out. Also like POPE, The selected MLLMs will also be prompted \textit{\textsl{Yes or No}}
   questions.

   \renewcommand{\arraystretch}{1.5}
   \begin{table*}[htbp]
     \centering
      \begin{tabular}{l l c l l l l l}
          \toprule
          setting & model & method & Accuracy&  Precision & Recall&  F1-Score & Yes Rate \\
          \midrule
          \multirow{5}{*}{adversarial} &\multirow{2}{*}{QWen}&  plain  & 0.8417 & 0.8970 &  0.772 & 0.8298& 0.4303 \\
          && +Piculet  & \textbf{0.8637} &\textbf{0.9225}&  \textbf{0.794}& \textbf{0.8535}&0.43033\\
          \cline{2-8}
          &\multirow{3}{*}{LLaVA} &  plain  & 0.7333 & 0.6902 & \textbf{0.8467} & 0.7605 & 0.6133 \\
          && +Woodpecker  & 0.8067&0.8286&  0.7733& 0.8000&0.4667\\
          && +Piculet  & \textbf{0.8597 } &\textbf{0.9397}&  0.7687 & \textbf{0.8456}&0.409 \\
          \hline
          \multirow{4}{*}{random} &\multirow{2}{*}{QWen} &  plain  &  0.8787 & 0.9805 &  0.7727 & 0.8643 & 0.394 \\
                                                             && +Piculet   & \textbf{0.8893}&\textbf{0.9811}&  \textbf{0.794}& \textbf{0.8777}&0.4047 \\
              \cline{2-8}
                                                            &\multirow{3}{*}{LLaVA}&  plain  & 0.8600 & 0.8750 &  \textbf{0.8400} & 0.8571 & 0.4800 \\
                                                            && +Woodpecker  & \textbf{0.8767} &0.9593&  0.7867& \textbf{0.8645}&0.4100\\
                                                            && +Piculet  & 0.8732  &\textbf{0.9813}&  0.7687 & 0.8621 & 0.4037\\
          \hline
          \multirow{4}{*}{popular} &\multirow{2}{*}{QWen} &  plain  &  0.8657 & 0.9492 &  0.7727 & 0.8519 & 0.407 \\
                                                            && +Piculet   & \textbf{0.8803} &\textbf{0.9597}&  \textbf{0.794}&\textbf{0.8690}& 0.4137 \\
                                                             \cline{2-8}
                                                            &\multirow{2}{*}{LLaVA}&  plain  & 0.7667 & 0.7222 &  \textbf{0.8667}& 0.7879 & 0.6000 \\
                                                            && +Woodpecker  & 0.8067&0.8382&  0.7600& 0.7972&0.4533\\
                                                            && +Piculet  & \textbf{0.87} &\textbf{0.9641}&  0.7687& \textbf{0.8553}&0.3987\\
          \bottomrule
      \end{tabular}
      \captionof{table}{Results on POPE using Qwen-VL-7B and LLaVa-v1.5-13B as baseline model.
      +Piculet denotes MLLM responses generated by our proposed Piculet, and +Woodpecker for woodpecker's method. The best performances within each setting are bolded. Our method achieves a near-universal advantage across the board.}
     \label{tab-pope}
   \end{table*}

   \textbf{LLaVA-QA90.} The LLaVA-QA90\cite{LLaVa-QA90} contains randomly selected 30 image for COCO-Val-2014, and for each
   image, three types of questions (conversation, detailed description, complex reasoning) are generated using the proposed data generation pipeline in \cite{LLaVa-QA90}.
  Specifically, we sample 10 description-type queries that are paraphrased in various forms to instruct an
  MLLM to describe an image, such as "\textit{\textsl{Analyze the image in a comprehensive and detailed manner.}}" and 
  "\textit{\textsl{Explain the visual content of the image in great detail.}}". 
  GPT-4V \cite{gpt4v} is utilized to evaluate the answers generated by the plain baseline model and our framework's model. We directly feed the image to GPT-4V, and prompt it to rate the responses regarding our designed two dimensions, i.e., accuracy and detailedness. The prompt template is available in Fig \ref{prompt-gpt4}.

  \renewcommand{\arraystretch}{1.5}
  \begin{table}[htbp]
    \centering
     \begin{tabular}{l c ccccccc}
         \toprule
          model& method & Total    &  Existence & Count&  Position & Color & Celebrity & OCR \\
         \midrule
         \multirow{2}{*}{QWen} &     plain  & 871.12 & 185 &  140 & \textbf{128.33} & 180  & 135.29&\textbf{102.5}\\
         &+Piculet  & \textbf{944.12} &\textbf{190.0}&  \textbf{163.34}& 126.67&\textbf{185.0}&\textbf{184.12}& 95.0\\
         \hline
       \multirow{3}{*}{LLaVA} &     plain  &698.72 & 195 &  95 & 53.33 &78.33 & 152.06& \textbf{125.0} \\
       &+Woodpecker  &- &195 &  \textbf{160.00}& 55.00&155.00&-&-\\
       &+Piculet  & \textbf{928.9} &185.0 &  151.66 &\textbf{121.66}&\textbf{175.0}&\textbf{195.58}&110.0\\
         \bottomrule
     \end{tabular}
     \captionof{table}{Results on MME using Qwen-VL-7B and LLaVa-v1.5-13B as baseline model. +Piculet denotes MLLM responses generated by our proposed Piculet, and +Woodpecker for woodpecker's method. The best performances within each setting are bolded. Our method achieves a near-universal advantage across the board.}
    \label{tab-mme}
  \end{table}
   
  \renewcommand{\arraystretch}{1.5}
  \begin{table}[htbp]
    \centering
     \begin{tabular}{l l cc}
         \toprule
          models&method &  accuracy    &   detailedness  \\
         \midrule
         \multirow{2}{*}{QWen} &plain  & 6.1 & 5.5  \\
       &+Piculet  & \textbf{7.3}&  \textbf{6.5}\\
          \hline
          \multirow{2}{*}{LLaVA} &plain  &5.6 & 5.9 \\
          &+Piculet  & \textbf{6.8}&  \textbf{6.3}\\
         \bottomrule
     \end{tabular}
     \captionof{table}{Results on LLaVa-QA90 using Qwen-VL-7B and LLaVa-v1.5-13B as baseline model. with denotes MLLM responses generated by our proposed Piculet. We don't know Woodpecker' exact 10 sampled examples, so cannot compare with their scores. The accuracy and detailedness metrics are on a scale of 10, and a higher score indicates better performance. The best performances within each setting are bolded. Our method achieves better performances on both accuracy and detailedness aspects.}
    \label{tab-llava}
  \end{table}
  
  \subsection{Experimental Results}
  \textbf{Resutls on POPE.} Instead of sampling several hundreds of images, 
  We directly utilized the entire dataset, which amounts to 9,000 image-text queries, thereby enabling a more thorough and comprehensive comparison to 
  demonstrate the superiority of our method. The tested results on POPE are shown in Table \ref{tab-pope}, which utilizes Qwen-VL-Chat\cite{qwenvl} and LLaVa\cite{LLaVa-QA90} as baseline model. 
  Considering that Woodpecker\cite{woodpecker} is the most similar training-free method to ours, we also compare with their tested results.
  As can be seen from the results, our 
  proposed framework achieves an across-the-board performance improvement on all test sets and in all aspects. In detail, in all the adversarial, random, and popular testset, our 
  method outperforms both the plain baseline model and the Woodpecker-enhanced model in all the accuracy, precision, recall and f1-score,
   except the only one random set, where our method is slightly inferior to Woodpecker. 

   A seemingly counterintuitive point is that the unenhanced, plain LLaVa actually performs the best on Recall, compared to both Woodpecker and our Piculet.
   However, this is actually reasonable: because MLLM inherently tends to answer "yes" to all \textit{\textsl{Yes or No}}
   questions, without discrimination. This results in Recall, a measurement that measures the proportion of correctly classified samples out 
   of all correct samples, being higher than that of both Woodpecker and our Piculet.
   The relatively high Yes Rate score of plain LLaVA also corroborates this speculation, which attests to our
    algorithm's effectiveness in mitigating the hallucinations of MLLMs as well.
   
Overall, our method outperforms Woodpecker, not to mention that our method operates with faster inference and lower resource consumption, while merely providing additional factual external knowledge to the MLLM, allowing the MLLM to make its own decisions and produce outputs based on the reliable information.

  \textbf{Resutls on MME.} The results on MME are shown in Table \ref{tab-mme}, and the baseline model is also Qwen-VL-Chat and LLaVa. For comparison, we conducted experiments on MME's existence, count, position, color, celebrity and ocr test set. For comparison, we also add the Woodpecker's results in the table. As can be seen from the table, our Piculet outperforms Woodpecker in most aspects, with only existence and count set as exceptions, where it slightly lags behind Woodpecker. Even so, considering that our method merely incorporates a few additional small deep learning models apart from the MLLMs to be corrected, and it significantly reduces inference time and operational costs compared to Woodpecker, our method is undoubtedly the better choice.

  \textbf{Resutls on LLaVA-QA90.} The results on LLaVA-QA90 is shown in Table \ref{tab-llava}, and the baseline model is also Qwen-VL-Chat and LLaVa. In this experiment, we
  sampled 10 description-type queries, which are paraphrased in various forms to instruct an MLLM to describe an image, to evaluate our proposed framework's 
  performance. The results, as can be seen from the table, show that our method has also achieved superior performance in both evaluation aspects. It's 
  worth noting that, as Woodpecker's sampled 10 queries are not exactly known, so we can't compare with their results here.
  
  \subsection{Ablation Study.}
We conduct an ablation study on MME datasets to validate the superiority and effectiveness of our method. In this section, we utilize Qwen-VL-Chat
as baseline model, in each test set, we select two of three specialized models and run experiments to compare the generated results' scores.
The calculated results are shown in Table \ref{tab-ablation-study}.
\renewcommand{\arraystretch}{1.5}
\begin{table}[htbp]
  \centering
   \begin{tabular}{l l lc lll l ll}
       \toprule
        Detection & OCR & Face  &Total&  Existence & Count&  Position & Color & Celebrity & OCR \\
       \midrule
      & \checkmark  &\checkmark &922.40&185.0& 140.0& 126.67 & 185.0 & 183.24 & \textbf{102.5} \\
      \checkmark  & &\checkmark &941.62& 190.0& 163.33& 126.67 & \textbf{190.0} &  184.12&87.5 \\
      \checkmark  & \checkmark& & 888.24&190.0& 163.33& 126.67 & 185.0 &  128.24 &  95.0\\
      \checkmark  & \checkmark&\checkmark &\textbf{944.12}& \textbf{190.0}& \textbf{163.33}& \textbf{126.67} & 185.0 &  \textbf{184.12} &95.0 \\
       \bottomrule
   \end{tabular}
   \captionof{table}{ Ablation study results on MME using Qwen-VL-7B as baseline model. \checkmark means results generated utilizing corresponding specialized models.}

  \label{tab-ablation-study}
\end{table}

As can be seen from the experimental results, each specialized model manages to boost the score on its respective test set. Specifically, the results with the specialized Detection model outperform those without it in both existence and count scores. Similarly, the use of the specialized OCR model leads to higher scores on  the OCR test set compared to when it is not used. The same can be said for the specialized Face model. Based on the comprehensive comparative experimental results, we can confidently say that each of our specialized models contributes to improved outcomes. That is, the method we propose, which we name Piculet, can mitigate the hallucination phenomena in MLLMs, making the responses to users' queries more authentic and reliable.

  \section{Conclusion}
  In this paper, we propose a novel framework named Piculet to address the hallucinations of MLLMs at its root. As a training-free method, our approach requires only single one inference of the target MLLM, and several other small deep-learning models, no other rather large-scale models are involved, which is economical and time-saving, and is plug-and-play in various different MLLMs. We have achieved the goal of reducing hallucinations by supplying the MLLMs with dependable external knowledge generated by specialized models. We evaluate our method on numerous datasets with other methods, and the results demonstrate the effectiveness and improvement of our method. We hope that our method can contribute a small improvement and offer some insights into the handling of hallucinations of MLLMs, thus inspiring further research and development in the field.\footnote{The authors have no competing interests to
  declare that are relevant to the content of this article.}.

\end{document}